\pdfoutput=1

\documentclass[11pt]{article}

\usepackage{EMNLP2023}

\usepackage{times}
\usepackage{latexsym}

\usepackage[T1]{fontenc}

\usepackage[utf8]{inputenc}

\usepackage{microtype}

\usepackage{inconsolata}

\usepackage{booktabs}       
\usepackage{amsfonts}       
\usepackage{nicefrac}       
\usepackage{microtype}      
\usepackage{times}
\usepackage{url}
\usepackage{latexsym}

\usepackage{lipsum}
\usepackage{graphicx}
\usepackage{multirow}
\usepackage{todonotes}
\usepackage{amsmath}
\title{Syntactic Knowledge via Graph Attention with BERT in Machine Translation}

\author{Yuqian Dai, Serge Sharoff, Marc de Kamps \\
        University of Leeds, LS2 9JT, United Kingdom \\ \texttt{\{mlyd,s.sharoff,m.dekamps\}@leeds.ac.uk}}

\begin{document}
\maketitle
\begin{abstract}
Although the Transformer model can effectively acquire context features via a self-attention mechanism, deeper syntactic knowledge is still not effectively modeled. To alleviate the above problem, we propose Syntactic knowledge via Graph attention with BERT (SGB) in Machine Translation (MT) scenarios. Graph Attention Network (GAT) and BERT jointly represent syntactic dependency feature as explicit knowledge of the source language to enrich source language representations and guide target language generation. Our experiments use gold syntax-annotation sentences and Quality Estimation (QE) model to obtain interpretability of translation quality improvement regarding syntactic knowledge without being limited to a BLEU score. Experiments show that the proposed SGB engines improve translation quality across the three MT tasks without sacrificing BLEU scores. We investigate what length of source sentences benefits the most and what dependencies are better identified by the SGB engines. We also find that learning of specific dependency relations by GAT can be reflected in the translation quality containing such relations and that syntax on the graph leads to new modeling of syntactic aspects of source sentences in the middle and bottom layers of BERT.
\end{abstract}

\section{Introduction}
Neural Machine Translation (NMT) performs better than traditional statistical machine translation models to produce more fluent results, but they overlook some syntax resulting in translations with syntactic errors. The proposed Transformer model \cite{vaswani2017attention} has a self-attention mechanism but still cannot avoid syntactically incorrect translations with a limited bilingual training set. Inspired by the Transformer model, \cite{devlin-etal-2019-bert} propose the pre-trained model BERT, which not only preserves the structure of the Transformer but also features pre-training, a process of unsupervised learning on a large-scale corpus in advance. Rich knowledge in BERT and robust model structure provide a better initialization and unified framework for downstream tasks. Therefore, BERT, which can be pre-trained for monolinguals and has rich implicit linguistic knowledge, has received attention on Machine Translation (MT) tasks \cite{zhu2020incorporating,yan2022boosting}.

Explicit linguistic knowledge, such as syntax, has been widely used to improve the performance of NMT models, resulting in smoother outputs. In linguistics, syntactic dependencies in sentences are not given as sequences. Although they can be processed and represented by sequential models such as RNN variants or Transformer model \cite{egea-gomez-etal-2021-syntax,peng2021syntax}, but linear representations do not accurately represent all syntactic structures and phenomena. The recently proposed Graph Attention Network (GAT) \cite{velivckovic2017graph} can represent syntactic structures and inter-word dependencies more explicitly through topology. Moreover, since the representation of knowledge is given explicitly, it has better readability and interpretability, drawing much interest in Natural Language Processing (NLP) \cite{huang2020syntax,li2022automatic}. Can explicit syntactic knowledge incorporation via GAT with implicit knowledge from BERT improve translation quality? If yes, which syntactic relations in the source language benefit the most when generating the target language? We still lack interpretability in linguistics to discuss the fusion of graph neural networks and pre-trained models in MT scenarios, however.

In response, we propose Syntactic knowledge via Graph attention with BERT (SGB) model. The model combines syntactic information from source sentences via GAT with BERT, aiming to improve Transformer-based NMT by applying syntax and pre-trained model BERT. We use the multi-head attention on the graph to explicitly utilize the source-side syntactic dependencies as syntactic guides to complement the source-side BERT and the target-side decoder. Our experiments containing Chinese (Zh), German (De), and Russian (Ru) to English (En) translation tasks are designed to demonstrate the effectiveness of our approach. Our main contributions are as follows.
\begin{itemize}
\item To the best of our knowledge, SGB is the first attempt to demonstrate the effectiveness of combining syntactic knowledge on graph attention and BERT in MT tasks. It can be fine-tuned to complete the training of the MT engine without the need for pre-training from scratch.
\item We explore before-and-after changes in translation quality in terms of syntactic knowledge and Quality Estimation (QE) score. Our models improve the translation quality of three MT tasks without sacrificing the BLEU score, with more fluent translations for short and medium-length source sentences. Additionally, our investigation for source sentences clarifies which dependency relations in the source sentence are learned more effectively by the model to produce better translations.
\item We investigate the interpretability of translation quality improvement in terms of syntactic knowledge. GAT learning of syntactic dependencies can be reflected in translation quality. Learning and representing syntactic relations via GAT leads to new modeling of the source sentence by the lower and middle layers of BERT. The syntactic knowledge on the graph and the features reconstructed by BERT are some of the reasons that cause the translation quality to change.
\end{itemize}

\section{Related Work}
In recent years, pre-trained models have received much attention in NLP, and Transformer is a typical model framework for them \cite{devlin-etal-2019-bert,liu2019roberta}. BERT is a representative pre-trained model that uses two pre-training objectives for self-supervised learning on a large corpus. Masked Language Model (MLM) uses context to predict the masked words in the sentence by the context content. Next Sentence Prediction (NSP) determines whether two sentences are next to each other. These two objectives allow BERT to learn large amounts of implicit linguistic knowledge through self-supervised learning, where such linguistic knowledge can also be applied to downstream tasks through fine-tuning. Given that BERT has mastered some linguistic knowledge, many researchers try using BERT as an encoder or decoder module in NMT to assist the MT model in sentence modeling and improve translation performance. \cite{imamura2019recycling} find that using BERT directly as an encoder in an MT system and employing two-stage optimization can improve low-resource language learning. \cite{yang2020towards} optimise BERT for catastrophic forgetting in MT tasks by using a concerted training framework. \cite{zhu2020incorporating} use the attention module to fuse the output features of BERT to the encoder and decoder to realize that the MT model can fully use the knowledge from BERT and self-adaptive learning.

Syntactic dependency is essential in MT tasks, aiming to analyze the grammatical structure of sentences and represent it as an easily understandable tree structure. Such explicit structural information helps the MT model understand sentence context information better and reduces sentence ambiguity. Several works have demonstrated the benefits of introducing syntactic information into NMT. \cite{currey-heafield-2019-incorporating} linearize and inject syntactic information of the source sentence into the Transformer model and discuss the performance gain in a low-resource translation task. \cite{zhang2020sg} introduce the syntactic dependency of interest into the attention mechanism and combine it with the Transformer model to obtain a better linguistics-inspired representation. \cite{mcdonald-chiang-2021-syntax} present different masks to guide the observation of attention mechanisms based on syntactic knowledge, and the attention head can select and learn from multiple masks. However, syntactic information is mostly modeled linearly, and syntactic knowledge of topological representations still lacks sufficient discussion. Furthermore, they discuss the application of syntactic knowledge in the Transformer model and the scenario when BERT in MT models is not investigated.

Graph neural networks can be regarded as a method of feature integration where the nodes represent the words in the sentence, and the edges describe the connections between the words. The definition of graph structure for a sentence is the key to designing a graph neural network, as it needs to be set in advance and cannot be changed during training. Therefore, the graph structure is a combination of prior knowledge and explicit features represented on the graph. Recently proposed GAT can efficiently represent data in non-euclidean spaces and combine attention mechanisms to assign weights to different nodes on the graph, independent of the specific network structure. Learning on graphs and supporting multi-headed attention mechanism, GAT can be used to represent linguistic knowledge in downstream tasks in combination with BERT \cite{huang2020syntax,chen2021combining,zhou2022dynamic}. Most studies only discuss syntactic knowledge and BERT singularly in MT scenarios, and whether explicit syntactic knowledge via GAT and BERT can improve translation quality in MT tasks is still being determined. Also, there is a lack of interpretability from the perspective of linguistic knowledge to bring more information about the changes in translation quality.

\section{Methodology}
\begin{figure*}[!ht]
\centering
\includegraphics[scale=0.6]{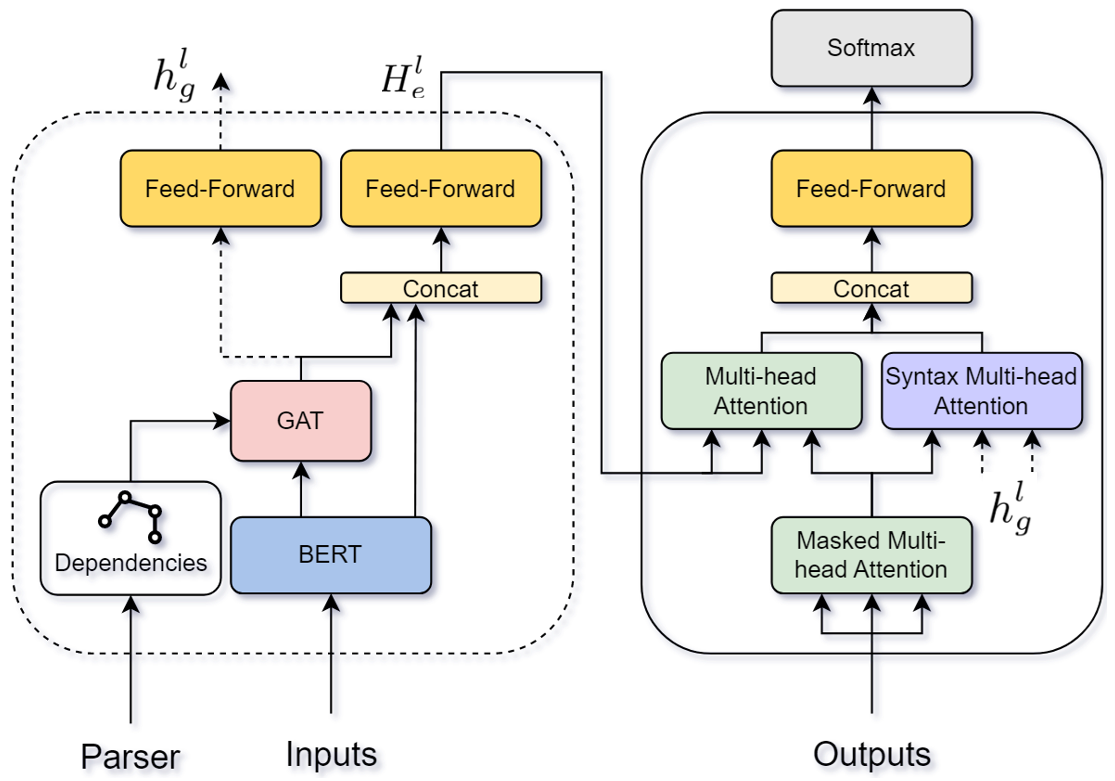}
\caption{The architecture of the SGB engines. The encoder with BERT and GAT on the left and the decoder on the right. Dash lines indicate the alternative connections. $H_{e}^{l}$ and $h_{g}^{l}$ represent the final layer output of BERT and GAT.}
\label{SGB_Model} 
\end{figure*}

In this section, we cover the descriptions of each of the layers in the engine. Figure \ref{SGB_Model} shows the overall architecture of the proposed SGB engine, which consists of encoding layer, graph attention layer, fusion and output layer.

\subsection{Encoding}
The experiments include translations from three source languages into English: Chinese to English (Zh$\rightarrow$En), Russian to English (Ru$\rightarrow$En), and German to English (De$\rightarrow$En). Given source sentence $S=[w_{1},w_{2},w_{3},\dotso w_{i}]$, where $i$ is the number of tokens in a sentence, $S$ is then cut into subword tokens and fed into BERT, which become: $\tilde{S} = {[[CLS],w_{1}^{1},w_{1}^{1\#1},w_{2},w_{3}^{3},w_{3}^{3\#3},\dotso w_{n},[SEP]]}$, Where $w^{n\#n}$ represents the subwords of $w_{n}$, [CLS] and [SEP] are special tokens of BERT. 

We use three BERT variants as an encoder for each MT engine, where Chinese is chinese-bert-wwm-ext\footnote{\url{https://huggingface.co/hfl/chinese-bert-wwm-ext}}, Russian is rubert-base\footnote{\url{https://huggingface.co/DeepPavlov/rubert-base-cased}}, and German is bert-base-german\footnote{\url{https://huggingface.co/bert-base-german-cased}}. Although their model structures are the same, the approaches differ in pre-training. Chinese BERT uses Whole Word Masking, Russian BERT takes the multilingual version of BERT-base as its initialization, and the approach of German remains the same as vanilla BERT. We aim to propose approaches that can be generalized to the BERT model structure, although their pre-training objectives are unique.

By capturing the representation of each subword token through BERT, the final embedded sequence is accessible via the last layer of BERT, $h_{B} = BERT(\tilde{S})$. To obtain the syntactic dependency information of the source sentence $\tilde{S}$, we use a Universal Dependencies-based parser\footnote{\url{https://github.com/hankcs/HanLP}} to perform tokenizing and syntactic dependency parsing on source sentences. 

After obtaining the parsing results, we construct the node adjacency matrix for graph representation. Each token has a corresponding node in the graph. Since word representation from BERT contains rich semantic information, nodes on the graph are encoded by BERT. Considering the subword segmentation, we merge subword token representation in an average way to obtain the node embeddings on the graph.

\subsection{Graph Attention}
Words and adjacency relations in a sentence can be represented as a graph structure, where the words on the graph are as nodes, and the relationships called syntactic dependencies between words are regarded as edges connecting nodes. We use GAT \cite{velivckovic2017graph} as our critical component to 
fuse the graph-structured information and node features. The node features given to a GAT layer are $\tilde{S}=[x_{1},x_{2},\dotso x_{i},\dotso x_{n}], x_{i}\in\mathbb{R}^{F}$, where $n$ is the total number of nodes, ${F}$ is the feature size of each node. The Equation (1) and (2) summarise the working mechanism of the GAT.
\begin{equation}
\small
h_{i}^{out}=\mathop{\parallel} \limits_{k=1}^K\sigma\left(\displaystyle\sum\limits_{j\in  N_{i}}\alpha_{ij}^{k}W^{k}x_{j}\right)
\end{equation}
\begin{equation}
\small
\alpha_{ij}^{k} = \frac{exp(LeakyReLU(a^{T}[Wx_{i} \parallel Wx_{j}]))}{\sum_{v\in N_{i}}exp(LeakyReLU(a^{T}[Wx_{i} \parallel Wx_{v}]))}
\end{equation}

1-hop neighbors $j\in N_{i}$ are attended by the node ${i}$, $\mathop{\parallel} \limits_{k=1}^K$ represents ${K}$ multi-head attention output concatenation. $h_{i}^{out}$ is the representation of node ${i}$ at the given layer. $\alpha_{ij}^{k}$ means attention between node ${i}$ and ${j}$. $W^{k}$ is linear transformation, $a$ is the weight vector for attention computation, $LeakyReLU$ is activation function. Simplistically, the feature calculation of one-layer GAT can be concluded as $h_{G}=GAT(X,A;\Theta^{l})$. The input is $X \in\mathbb{R}^{n \times F}$, and the final output is $h_{G} \in\mathbb{R}^{n \times F'}$ where n is the number of nodes, $F$ is the feature size for each node, $F'$ is the hidden state for GAT, $A\in\mathbb{R}^{n \times n}$ is the graph adjacency matrix indicating node connection, $\Theta^{l}$ is the parameters during training.

\subsection{Fusion and Output}
We propose two approaches for using syntactic knowledge in MT engines. The first, called Syntactic knowledge via Graph attention with BERT Concatenation (SGBC), is to combine the syntactic knowledge on the graph and BERT to work on the encoder, as shown in Equations (3) and (4). 
\begin{equation}
\small
H_{e}^{l}=concat(h_{B},h_{G})
\end{equation}
\begin{equation}
\small
\tilde h_{d}^{l} = attn_{D}(h_{d}^{l},H_{e}^{l},H_{e}^{l})
\end{equation}
where $attn_{D}$ stands for encoder-decoder attention in MT engines. $l$ is the output of the $l$-th layer, $d$ is the representation of the tokens in decoder-side. $H_{e}^{l}$ contains the features of BERT ($h_{B}$) and GAT ($h_{G}$) fed into the encoder-decoder attention module in the decoder. The feed-forward network subsequently processes the attention features alone with residual connection, as in the case of the vanilla Transformer model.

The second one, called Syntactic knowledge via Graph attention with BERT and Decoder (SGBD), is that the syntactic knowledge on the graph is not only applied to the encoder but also guides the decoder through the syntax-decoder attention, as shown in Equations (5), (6) and (7).
\begin{equation}
\small
\tilde h_{d}^{l} = attn_{D}(h_{d}^{l},H_{e}^{l},H_{e}^{l})
\end{equation}
\begin{equation}
\small
\tilde h_{s}^{l} = attn_{S}(h_{d}^{l},h_{g}^{l},h_{g}^{l})
\end{equation}
\begin{equation}
\small
\tilde h_{t}^{l} = concat(\tilde h_{d}^{l},\tilde h_{s}^{l})
\end{equation}
where $attn_{D}$ and $attn_{S}$ represent encoder-decoder attention and syntax-decoder attention respectively. $h_{g}^{l}$ is the output of GAT containing syntactic dependency features of sentences via another feed-forward network. $\tilde h_{t}^{l}$ is the final attention features obtained by concatenating $attn_{D}$ and $attn_{S}$. As with the vanilla Transformer, the predicted word is generated by a feed-forward network with residual connection and softmax function.

\section{What happens to model performance?}
We evaluate the effectiveness of the proposed approach by BLEU score on the UNPC\footnote{\url{https://opus.nlpl.eu/UNPC.php}} and Europarl\footnote{\url{https://opus.nlpl.eu/Europarl.php}} datasets, which are UNPC Chinese-English (Zh$\rightarrow$En) and Russian-English (Ru$\rightarrow$En), and Europarl German-English (De$\rightarrow$En), respectively. We select 1M sentence pairs as the training set for each language and 6K and 5K sentence pairs as the validation and test sets. In addition, we progressively reduce the training set so that it can simulate the effect of our approach on other low-resource languages and limited training set scenarios. 

Its MT engine for the encoder is a single BERT, which is the baseline model (Baseline). The baseline and our proposed SGB engines are consistent regarding model training for a fair comparison. The decoders are from the vanilla Transformer model, except for BERT variants for each source language. They have 6 layers and 8 attention heads, while other parameters are kept consistent. The GAT in SGB engines has 2 layers and 6 attention heads for Zh, 4 attention heads for Ru and De. The MT engines are trained using the Adam optimizer with parameters $\beta_1 = 0.9$ and $\beta_2 = 0.98$. The learning rate is 2e-5, word embedding = 768, and the cross entropy as loss function. All experiments are performed on RTX 3080 and 3090 GPUs.

\begin{table}[!ht]
\centering
\small
\begin{tabular}{c|lccc} 
\hline
Data set size   & Zh$\rightarrow$En   & Baseline  & SGBC  & SGBD   \\ 
\hline
\multirow{8}{*}{1M} & BLEU & 47.15 & \textbf{47.23} & 47.17  \\
                    & COMET   & 82.20     & 83.69     & \textbf{84.78}      \\ 
\cline{2-5}
                    & Ru$\rightarrow$En   & Baseline  & SGBC  & SGBD   \\ 
\cline{2-5}
                    & BLEU & 47.22 & \textbf{47.36} & 47.27  \\
                    & COMET   & 80.93     & 81.34     & \textbf{82.56}      \\ 
\cline{2-5}
                    & De$\rightarrow$En   & Baseline  & SGBC  & SGBD   \\ 
\cline{2-5}
                    & BLEU & 37.59 & \textbf{37.67} & 37.63  \\
                    & COMET   & 78.02     & 78.66     & \textbf{79.37}      \\
\hline
\end{tabular}
\caption{BLEU and QE scores for translations of three languages under 1M training set size.}
\label{OverallResults} 
\end{table}

As shown in Table \ref{OverallResults}, the proposed SGB engines perform well under different source languages, achieving comparable or better BLEU scores than the baseline models. The SGB engines also show improvement compared with the baseline in the case of small training samples, which may also improve the performance of other low-resource languages or limited training set scenarios (see Appendix Sec \ref{BLEU for Translations} for details). Explicit syntactic knowledge represented by graph attention and BERT is beneficial to learning linguistic structure by the MT models. Inspired by \cite{kocmi-etal-2021-ship}, we also use the COMET QE model to re-evaluate the performance of the engines, where COMET gives a QE score from 0 to 100, considering the relationship between the source sentence, the translation, and the reference. We find that the SGB engines all have higher BLEU and QE scores. But in actual translation tasks where a reference is missing, and there is a need for translation both in and out of the domain, the QE model is a better metric than BLEU for addressing such an urgent situation.

\section{What happens to translation quality?}
After applying our approach in the MT task, we investigate how the specific translation quality is improved. Given that the BLEU score does not reflect linguistic information of the sentence or human judgment \cite{callison-burch-etal-2006-evaluating,novikova2017we}. Therefore, we use the gold syntactic annotation corpus and the QE model, which concerns factors such as the retention of source sentence semantics, the coherence of translation semantics, and the rationality of the word order in the translation to explore the changes in translation quality and the effectiveness of our approach in terms of syntactic knowledge.

\subsection{Overall Translation Quality}
We translate the PUD corpus (PUD Chinese\footnote{\url{https://github.com/UniversalDependencies/UD_Chinese-PUD}}, PUD Russian\footnote{\url{https://github.com/UniversalDependencies/UD_Russian-PUD}}, and PUD German\footnote{\url{https://github.com/UniversalDependencies/UD_German-PUD}}) using the baseline and SGB engines for three languages. All 1,000 sentences in each PUD corpus for each language are arranged in the same sequence and have the same meaning. We then use the state-of-the-art QE model\footnote{\url{https://github.com/TharinduDR/TransQuest}} to score these translations from 0 to 1, with a higher score representing better translation quality. We use a paired t-test and box plot to investigate the changes and distribution in translation quality before and after our approaches, where the significant level = 0.05 in paired t-test.

From Table \ref{P_t_test}, when comparing the Zh baseline and SGBC models, $\bar{x}_{d}$ of them is 0.024, $S_{d}$ is 0.109 and the test statistic (t) is 7.18, corresponding to a p-value  $<$ 0.001. The t and p-value in SGBD also reveal the statistical significance of the QE scores before and after our approach. Both reject $H_0$ at the significant level of 0.05 ($H_0$ is that our approaches do not significantly differ in QE scores compared to the baselines.). Instead, $H_1$ is accepted that the differences between baseline and SGB engines in QE scores are large enough to be statistically significant. Similar results are observed for Ru and De, where the translation quality after applying our approach are significantly different than before via QE scores. The box plot also shows the QE score distribution for the baseline and two SGB engines for the three languages (see Appendix Sec \ref{Box Plot in Overall Translation Quality} for details). Syntactic knowledge via graph representation and BERT do improve the translation quality of the MT engines. SGBD engines receive higher QE scores, although BLEU prefers the SGBC model and scores higher.

\begin{table*}[!ht]
\centering
\small
\begin{tabular}{c|c|cc|cccc} 
\hline
\multicolumn{1}{l|}{Language}  &Sample size &\multicolumn{2}{c|}{Models}  & $\bar{x}_{d}$     & $S_{d}$     & t     & P-value  \\ 
\hline
\multirow{2}{*}{Zh}         & \multirow{2}{*}{1000}  & \multirow{2}{*}{Baseline} & SGBC & 0.024 & 0.109 & 7.18  & p $<$ 0.001 \\
                               &                      & & SGBD & 0.032 & 0.111 & 9.12  & p $<$ 0.001   \\ 
\hline
\multirow{2}{*}{Ru}           & \multirow{2}{*}{1000}  & \multirow{2}{*}{Baseline} & SGBC & 0.024 & 0.042 & 18.38 & p $<$ 0.001   \\
                               &                      & & SGBD & 0.034 & 0.045 & 23.67 & p $<$ 0.001  \\ 
\hline
\multirow{2}{*}{De}           & \multirow{2}{*}{1000} & \multirow{2}{*}{Baseline} & SGBC & 0.007 & 0.113 & 2.162 & p $=$ 0.030   \\
                               &                     &  & SGBD & 0.012 & 0.110 & 3.617 & p $<$ 0.001   \\
\hline
\end{tabular}
\caption{Paired t-test for PUD corpus translations of three languages between base and SGB models.}
\label{P_t_test} 
\end{table*}

\subsection{Sentence Length}
We also investigate the association between the proposed approaches and the sentence length of the source language. After translating the PUD corpus for the three languages using the baseline engines and scoring them using the QE model, we rank translations according to their QE scores from highest to lowest and consider the bottom 30\% of translations as low-quality translations. We divide low-quality translations again according to their source sentence length. Given that the sentence length of a source sentence is $x$, it is considered a short sentence (S) if $x \leq 25$. It is a medium sentence (M) if $25 < x \leq 45$. It is a long sentence (L) if $45 < x$. Considering the differences in characters and words in these three languages, both Russian and German follow another rule. The length of a source sentence is $x$, it is a short sentence (S) if $x \leq 14$. It is a medium sentence (M) if $14 < x \leq 24$. It is a long sentence (L) if $24 < x$. We then compare the average QE scores of these low-quality translations in the SGB engines with those in baseline engines to analyze which lengths of source sentences benefit the most.

Table \ref{Sen_L} shows that the translation quality of the proposed SGB engines is improved in all MT tasks. The SGBC engine is more effective for long source sentences, whereas the SGBD engine focuses more on improving short and medium-length source sentences. The translation quality improvement is more significant when short and medium-length source sentences come. But such an advantage is not reflected in BLEU, which can reflect the performance via one standard reference within the domain. However, the PUD corpus contains out-of-domain sentences (not only news but also wiki), which places higher demands on the generalization ability of the model and its ability to understand sentence structure. SGBD can achieve better translation results for out-of-domain sentences without sacrificing the BLEU performance in three languages MT tasks, which reflects that the graph syntactic information can enrich the source language representation and enhance the model ability to learn linguistic information.

\begin{table}[!ht]
\centering
\small
\begin{tabular}{ccccc} 
\hline
\multicolumn{5}{c}{Zh}                              \\
Sen L & Samples & Baseline  & SGBC   & SGBD \\ 
\hline
L     & 93       & 0.425 & \textbf{0.512} & 0.508            \\
M     & 142      & 0.423 & 0.500 & \textbf{0.517}           \\
S     & 65       & 0.434 & 0.543 & \textbf{0.560}            \\ 
\hline
\multicolumn{5}{c}{Ru}                              \\
Sen L & Samples & Baseline  & SGBC   & SGBD  \\ 
\hline
L     &   32      &   0.719    & \textbf{0.751}      &0.745                 \\
M     &   155       & 0.698      &  0.746     &    \textbf{0.750}              \\
S     &     113     &   0.686    & \textbf{ 0.752}     &    0.747              \\ 
\hline
\multicolumn{5}{c}{De}                              \\
Sen L & Samples & Baseline  & SGBC   & SGBD  \\ 
\hline
L     &      57    & 0.513      &    \textbf{0.554}   &    0.549              \\
M     &  150        &    0.512   &  0.561   &   \textbf{0.586}               \\
S     &     93     & 0.482      &    0.574   &    \textbf{0.578}             \\
\hline
\end{tabular}
\caption{QE scores of Baseline and SGB models for low-quality translations of different sentence lengths.}
\label{Sen_L} 
\end{table}

\subsection{Syntactic Relations}
Several different types of dependency relations indicate the structure of one given sentence. If our approaches help improve translation quality, which dependency relation in the source sentence benefits the most? We keep the low-quality translations and classify the sentence types according to dependency relations. Given a dependency relation $d$, source sentences of low-quality translations containing $d$ are grouped. We calculate the average QE score of them with this dependency relation before and after applying our approaches.

Table \ref{SyntaxRelations} presents the most improved syntactic relations for each language. The translation quality of source sentences classified according to dependency relations is improved to different degrees for each language under the proposed approaches (details are in Appendix Sec \ref{Sentence Relations Test}). Although SGBC and SGBD are both equipped with graph syntactic knowledge, their learning of dependencies is different, e.g., "flat" in Zh is significant in the SGBC but not in SGBD. SGBD, decoders also guided by syntactic knowledge on the graph, does not handle all the syntactic relations to have a higher QE score than the SGBC model, e.g., "discourse:sp", "orphan" and "csubj" scores in Zh, Ru and De are higher in SGBC. It may be the engine focusing too much on syntactic knowledge, which leads to a knowledge redundancy that impairs the translation quality. However, the significance of some relations is similar, they all happen in SGBC and SGBD engines, which means that syntactic knowledge via graph attention with BERT allows the MT engine to be more explicit about some common specific relations regardless of changes in approach.

\begin{table*}[!ht]
\centering
\small
\begin{tabular}{lcc|lcc} 
\hline
\multicolumn{6}{c}{\textbf{Zh}}                                                                   \\
             & Baseline  & \multicolumn{1}{l}{SGBC} &              & Baseline  & SGBD  \\ 
\hline
obl:agent    & 0.379 & 0.576                   & obl:agent    & 0.379 & 0.597            \\
discourse:sp & 0.388 & 0.502                   & iobj         & 0.387 & 0.511            \\
flat         & 0.387 & 0.494                   & nsubj:pass   & 0.423 & 0.545            \\
flat:name    & 0.415 & 0.518                   & appos        & 0.404 & 0.518            \\
mark:prt     & 0.435 & 0.532                   & discourse:sp & 0.388 & 0.501            \\ 
\hline
\multicolumn{6}{c}{\textbf{Ru}}                                                                   \\
             & Baseline  & \multicolumn{1}{l}{SGBC} &              & Baseline  & SGBD  \\ 
\hline
orphan      & 0.608 & 0.768                   & orphan      & 0.608 & 0.719            \\
aux          & 0.700   & 0.764                   & aux          & 0.700   & 0.777            \\
ccomp        & 0.681 & 0.745                   & ccomp        & 0.681 & 0.747            \\
flat:name    & 0.703 & 0.761                   & discourse    & 0.614 & 0.676            \\
fixed        & 0.688 & 0.742                   & fixed        & 0.688 & 0.750             \\ 
\hline
\multicolumn{6}{c}{\textbf{De}}                                                                   \\
             & Baseline  & \multicolumn{1}{l}{SGBC} &              & Baseline  & SGBD \\ 
\hline
csubj        & 0.449 & 0.566                   & flat         & 0.442 & 0.625            \\
flat         & 0.442 & 0.553                   & csubj        & 0.449 & 0.554            \\
expl         & 0.486 & 0.573                   & expl         & 0.486 & 0.589            \\
compound:prt & 0.493 & 0.579                   & compound:prt & 0.493 & 0.595            \\
compound     & 0.495 & 0.577                   & cop          & 0.502 & 0.586            \\
\hline
\end{tabular}
\caption{Top-5 dependencies in source sentences where QE scores differ the most for each language.}
\label{SyntaxRelations} 
\end{table*}

\section{What happens to syntactic features?}
The explicit syntactic knowledge on the graph is beneficial for translation quality, but how GAT is linked to translation quality and influences the decisions of BERT is interesting to us. Therefore, we explore the interpretability of our approaches regarding syntax by performing syntactic prediction tests on GAT and representation similarity analysis with BERT.

\subsection{Syntactic Predictions in GAT}
One of the clues for improving translation quality is whether GAT has a syntactic understanding. What syntactic knowledge is straightforward for GAT to learn? To investigate whether there is a correlation between syntactic knowledge on graphs and translation quality, we design a syntactic dependency prediction task for GAT to investigate how it represents syntactic knowledge (see Appendix Sec \ref{Syntactic Predictions in GAT}). We use the PUD corpus as the training, validation, and test sets for each language, divided into 800, 100, and 100 sentences, respectively. The words and syntactic dependencies in the sentences are regarded as nodes and edges of the graph. GAT needs to learn the associations between nodes to predict different dependency relations and the evaluation metric is F1-score.

As shown in Table \ref{SyntaxPredGat}, the training cost of GAT is not expensive and only 2 layers are needed to achieve learning of dependency relations. By comparing the prediction score of dependency relations by GAT with source sentences containing such relation and their translation quality (see Appendix Sec \ref{Sentence Relations Test}), we find a link: the learning of dependency relations by the GAT can be reflected in the translation quality. E.g., the better performance in predicting 'conj' in Zh leads to a corresponding improvement in the translation of the source language containing this relation. We can find similar cases in Ru and De. However, some are difficult for GAT to predict, e.g., 'iobj' and 'nusbj:pass' both fail, and both 'obl:tmod' in Zh and De have a lower prediction score, but translation quality improves. (detailed results are in Appendix Sec \ref{Sentence Relations Test} and \ref{Syntactic Predictions in GAT}). One of the factors contributing to the improvement of translation quality can be the robust dependency relation learning by GAT. But it is not absolute since GAT may not effectively learn such features with fewer samples in the test, or the encoder or decoder needs a more explicit sentence structure information provided by GAT rather than whether the syntactic annotation is correct.

\begin{table*}
\centering
\small
\begin{tabular}{l|cr|l|cr|l|cr} 
\hline
\multicolumn{3}{c|}{\textbf{Zh}}                & \multicolumn{3}{c|}{\textbf{Ru}}                & \multicolumn{3}{c}{\textbf{De}}                  \\
\multicolumn{1}{l}{} & Samples & Score & \multicolumn{1}{l}{} & Samples & Score & \multicolumn{1}{l}{} & Samples & Score  \\ 
\hline
mark                 & 291     & 0.986 & det                  & 476     & 0.990  & case                 & 2053    & 0.992  \\
cc                   & 283     & 0.984 & root                 & 1000    & 0.987 & cc                   & 724     & 0.987  \\
conj                 & 383     & 0.970  & amod                 & 1791    & 0.982 & det                  & 2771    & 0.987  \\
nummod               & 809     & 0.965 & case                 & 2121    & 0.978 & mark                 & 459     & 0.981  \\
root                 & 1000    & 0.955 & aux:pass             & 128     & 0.974 & advmod               & 1103    & 0.932  \\
cop                  & 251     & 0.945 & cop                  & 87      & 0.971 & root                 & 1000    & 0.931  \\
det                  & 338     & 0.935 & advmod               & 914     & 0.934 & aux:pass             & 230     & 0.927  \\
case                 & 1319    & 0.934 & cc                   & 599     & 0.930  & amod                 & 1089    & 0.913  \\
nmod                 & 702     & 0.933 & flat:foreign         & 97      & 0.921 & flat:name            & 164     & 0.876  \\
amod                 & 420     & 0.927 & obl                  & 1465    & 0.900   & aux                  & 365     & 0.868  \\
\hline
\end{tabular}
\caption{Top-10 highest F1-sore for predicting dependency relations by GAT for each language.}
\label{SyntaxPredGat} 
\end{table*}

\renewcommand{\thefootnote}{\fnsymbol{footnote}}
\begin{table}
\centering
\small
\begin{tabular}{l|c|cc|cc} 
\hline
\multicolumn{6}{c}{\textbf{Zh}}                                                                              \\
\multicolumn{1}{c}{} & \multicolumn{1}{c}{GAT} & RSA   & \multicolumn{1}{c}{Layer} & RSA\footnote[1]   & Layer  \\ 
\hline
mark                 & 0.986                   & 0.178 & 4                         & 0.208 & 4      \\
cc                   & 0.984                   & 0.274 & 4                         & 0.354 & 5      \\
conj                 & 0.970                    & 0.380  & 5                         & 0.152 & 5      \\
nummod               & 0.965                   & 0.274 & 4                         & 0.237 & 3      \\
root                 & 0.955                   & 0.216 & 4                         & 0.390  & 4      \\ 
\hline
\multicolumn{6}{c}{\textbf{Ru}}                                                                              \\
\multicolumn{1}{c}{} & \multicolumn{1}{c}{GAT} & RSA   & \multicolumn{1}{c}{Layer} & RSA\footnote[1]   & Layer  \\ 
\hline
det                  & 0.990                    & 0.426 & 4                         & 0.408 & 3      \\
root                 & 0.987                   & 0.466 & 3                         & 0.504 & 3      \\
amod                 & 0.982                   & 0.444 & 3                         & 0.391 & 4      \\
case                 & 0.978                   & 0.462 & 4                         & 0.413 & 4      \\
aux:pass             & 0.974                   & 0.357 & 3                         & 0.327 & 3      \\
\hline
\multicolumn{6}{c}{\textbf{Ru}}                                                                              \\
\multicolumn{1}{c}{} & \multicolumn{1}{c}{GAT} & RSA   & \multicolumn{1}{c}{Layer} & RSA\footnote[1]   & Layer  \\ 
\hline
case                 & 0.992                    & 0.686 & 5                         & 0.759 & 2      \\
cc                & 0.987                   & 0.591 & 6                         & 0.741 & 6      \\
det                 & 0.987                   & 0.584 & 8                         & 0.817 & 6      \\
mark                 & 0.981                 & 0.676 & 6                         & 0.769 & 6      \\
advmod             & 0.932                  & 0.733 & 6                         & 0.774 & 8      \\
\hline
\end{tabular}
\caption{Top-5 highest F1-score of syntactic knowledge learning on the graph and its BERT layer with the lowest similarity in RSA analysis for each language.}
\label{RSA} 
\end{table}
\footnotetext[1]{Representations from Baseline and SGBD for comparison.}

\subsection{Representational Similarity Analysis}
Representational Similarity Analysis (RSA) is a technique used to analyze the similarity between different representation spaces of neural networks. Inspired by \cite{merchant2020happens}, RSA uses $n$ examples for building two sets of comparable representations between neural networks. The representations are then transformed into a similarity matrix and the Pearson correlation between the upper triangles of the similarity matrix is used to obtain the final similarity score between the representation spaces. We want to know whether the addition of syntactic knowledge on the graph also impacts the representation space of BERT and thus improves the modeling of source sentences. We divide the source sentences corresponding to the 300 low-quality translations according to the type of dependency relations as our stimulus. Given the current dependency relation is $x$, the source sentences of low-quality translations containing $x$ are all composed into one group stimulus. We extract BERT representations from both models for comparison (e.g., Baseline vs SGBC), and cosine similarity is used as the kernel for all experiments.

Table \ref{RSA} shows the results of our RSA analysis, BERT in the baseline, and SGB engines are compared based on syntactic prediction scores by GAT (details are in Appendix Sec \ref{Representational Similarity Analysis}). In all three languages, we observed that the lowest RSA scores usually occurred at the lower and middle layers of BERT. E.g., the lowest RSA scores are concentrated at layers 3-5 for Zh and Ru and layers 5-8 for German. Syntactic knowledge on the graph causes a sharp decrease in similarity in particular layers, suggesting that the process of MT fine-tuning involves refactoring syntactic knowledge, and the modeling of shallow and syntactic knowledge is more likely to change in BERT. Layers 9-12 tend to deal with higher-level semantic information. However, their similarity still differs, indicating that changes in the lower and middle layers also affect learning deep linguistic information in higher layers, although the last layer is task-oriented. It reveals that incorporating linguistic knowledge into the fine-tuned representation of BERT can lead to reconsidering such knowledge to obtain a more accurate representation of the source sentences and thus improve translation quality.

\section{Conclusions}
This paper proposes two approaches incorporating syntactic knowledge via GAT and BERT into the MT tasks. The experiments explain how and why translation quality is improved from the perspective of syntactic knowledge. In future work, we will continue to investigate how to model critical linguistic knowledge via graphs in MT tasks to improve translation quality.

\section{Limitations}
The corpus with gold syntactic annotations is expensive, containing only 1,000 syntax-annotated sentences for each language. If the PUD corpus could provide more annotated sentences, it would provide more accurate interpretability regarding syntactic knowledge. E.g., We can investigate more accurately how translation quality is being improved and explore more details than 1,000 sentences. For the prediction of syntactic dependency tasks on the GAT, we could avoid the phenomenon of the number of dependency relations being too small for experimental conclusions and know more accurately how the GAT learns syntactic knowledge. Also, this study uses an external parser to obtain the syntactic structure of the source language. The limitation is that the parser is also built based on neural networks and inevitably suffers from errors in syntactic knowledge annotation, which limits the syntactic knowledge on the graph to guide BERT and translation quality.

\bibliography{anthology,custom}
\bibliographystyle{acl_natbib}

\appendix
\clearpage
\section{Appendix}
\label{sec:appendix}

\subsection{BLEU and Size of the Training Set}
\label{BLEU for Translations}
The size of the training set is gradually reduced to investigate the performance of the SGB models, as shown in Table \ref{BLEU}, with BLEU serving as a generic and explicit performance metric.

\begin{table}[!ht]
\centering
\small
\begin{tabular}{ccccc}
\hline
                    & Size & Baseline & SGBC & SGBD \\ 
\hline
\multirow{3}{*}{Zh$\rightarrow$En} & 0.1M  &  24.26   & 24.89& 24.72             \\
                    & 0.5M  &  38.48    &38.71    &     38.53              \\
                    & 1M   & 47.15     &47.23     &  47.17                 \\ 
\hline
\multirow{3}{*}{Ru$\rightarrow$En} & 0.1M  &  21.12    & 21.45    &       21.33            \\
                    & 0.5M  &  37.69    &   37.74  & 37.68                \\
                    & 1M    & 47.22     & 47.36    &      47.27             \\ 
\hline
\multirow{3}{*}{De$\rightarrow$En} & 0.1M  & 15.41     &  15.79   &        15.50           \\
                    & 0.5M  &  26.89    &   27.13  &   26.92                \\
                    & 1M    &  37.59    &  37.67  &    37.63               \\
\hline
\end{tabular}
\caption{BLEU for translations of three languages with different training set sizes. Despite the smaller data set size, the SGB models are still more competitive than the Baseline model in terms of BLEU.}
\label{BLEU} 
\end{table}

\subsection{Box Plot for QE Scores in Overall Translation Quality}
\label{Box Plot in Overall Translation Quality}
\begin{figure*}
\centering
\small
\includegraphics[scale=0.24]{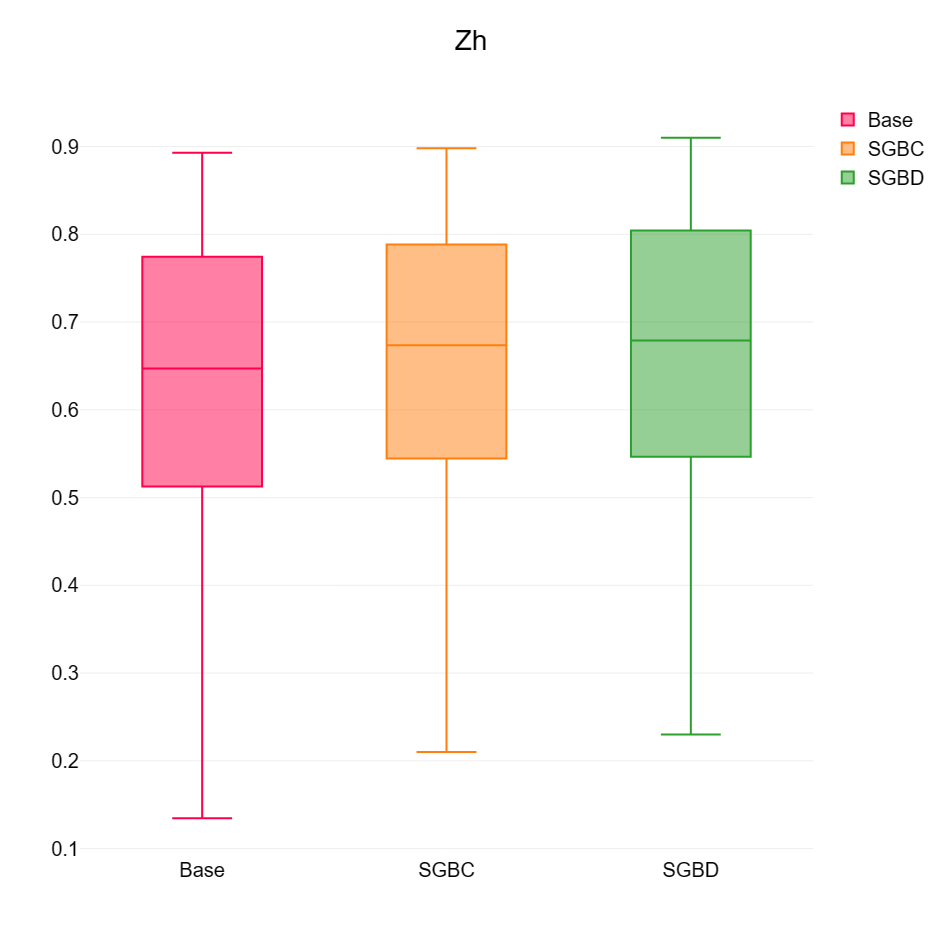}
\includegraphics[scale=0.24]{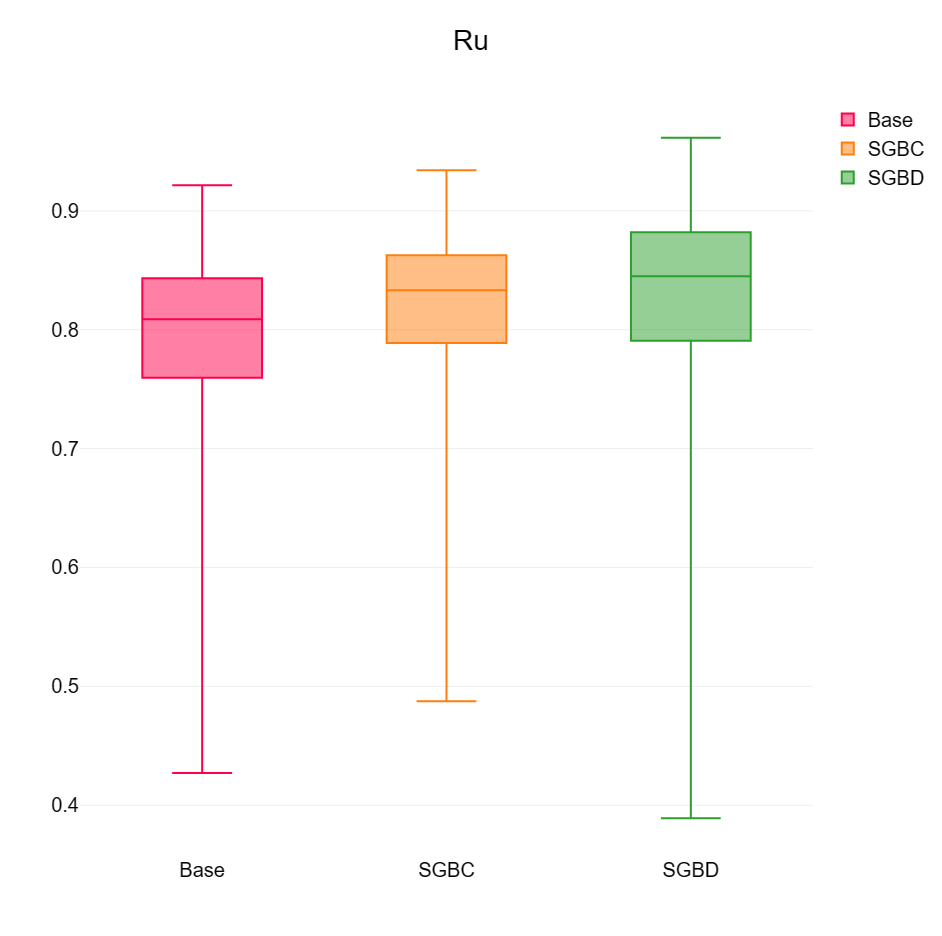}
\includegraphics[scale=0.24]{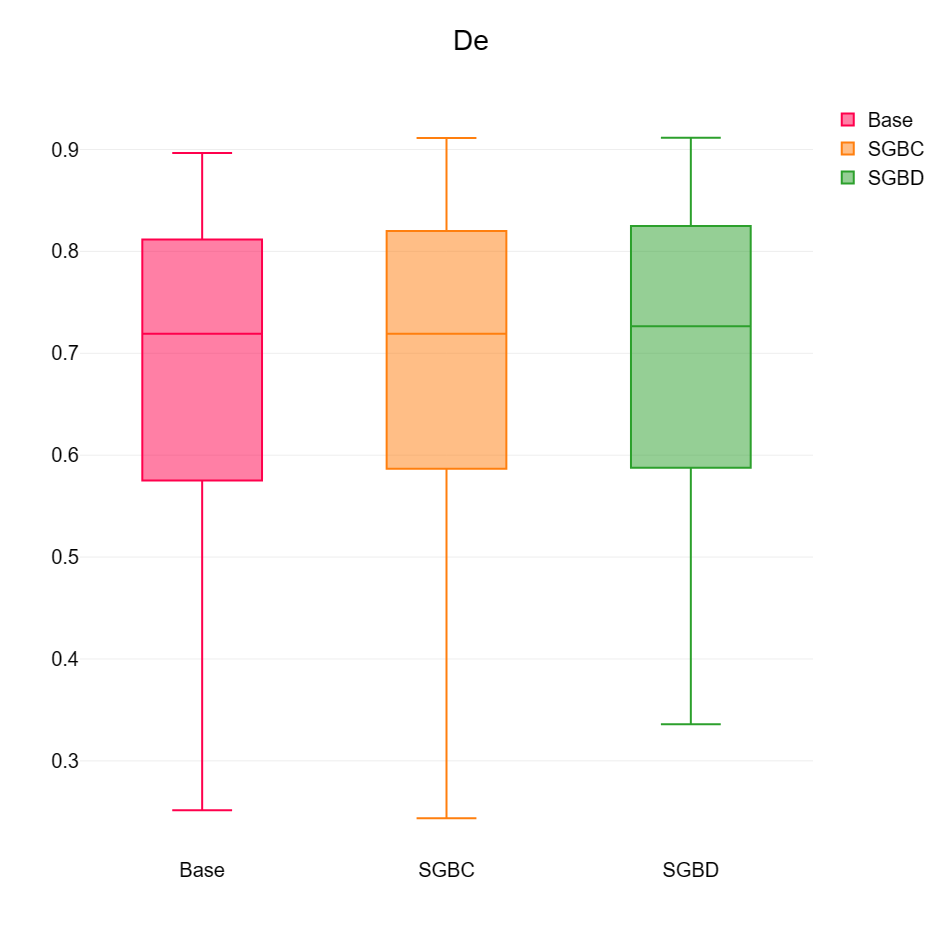}
\caption{The QE scores for the translations in the three languages are shown in the box plot. }
\label{boxplot} 
\end{figure*}

The QE scores for the three different translations from the source language into the target language English are presented in a box plot, as shown in Figure \ref{boxplot}.

\subsection{Sentence Relations Test}
\label{Sentence Relations Test}
Given the low-quality translations in each language, their corresponding source sentences are grouped according to different dependency relation features. Table \ref{FullZhSyntaxRelations} to Table \ref{FullDeSyntaxRelations} show the average QE scores of the baseline, SGBC and SGBD models for these groups of translations respectively, with higher scores associated with better translation quality. We retain some dependency relations, although they are few in number.

\subsection{Syntactic Predictions in GAT}
\label{Syntactic Predictions in GAT}
We design a syntactic dependency prediction task for GAT based on sentences annotated with gold syntax from the PUD corpus of three languages, where words in a sentence are considered nodes on a graph and dependency connections between words are considered as edges between nodes. GAT needs to predict edges type (dependency relations) based on node information (words). The dependency connections are treated as undirected graphs requiring the current node to consider information from all its neighbors.

In order to demonstrate the GAT mastery of syntactic information, the GAT that has yet to be trained for the MT task served as the test subject. The syntactic knowledge received by GAT in the MT task comes from the parser, which does not always provide the correct syntactic annotations. But it is still possible for GAT to model and represent them well in the MT task. If the syntactic dependencies of the gold annotation are used to test the MT-trained GAT, the prediction failure of the GAT does not indicate that it does not know the syntactic dependencies, given that the gold annotation and the parser annotation do not precisely match. Additionally, if the parser annotation is used to test the MT-trained GAT, although the GAT is good at predicting syntactic knowledge in the experiment, the annotation of this knowledge is wrong compared with the gold annotation. It does not reflect the true learning ability of the GAT since those knowledge does not correspond to linguistics.

In the experiments, the number of layers in GAT is 2. However, the number of attentional heads in GAT is 4 for Zh and 6 for all other languages, which is consistent with the parameters in the SGB model. Word embedding = 768, dropout = 0.2, optimiser = Adam, learning rate = 2e-5. Table \ref{GATSyntaxPred} shows the predictions of GAT for each language dependency relation. Given that some dependency relations are insufficient in the dataset, we remove them to ensure the accuracy of the experiment. "-" means that this language does not contain given dependency relation in the test.

\subsection{Representational Similarity Analysis}
\label{Representational Similarity Analysis}
Table \ref{Zh-RSA} to  Table \ref{De-RSA-2} show the RSA tests of the dependency relations in the given groups of BERT in the Baseline, SGBC and SGBD models for different languages in 12 layers (L).

\begin{table*}
\centering
\small
\begin{tabular}{l|c|ccc} 
\hline
\multicolumn{1}{c}{Bad Translations (Zh)} & \multicolumn{1}{c}{Sentences} & Baseline & SGBC   & SGBD  \\ 
\hline
acl:relcl              & 112                         & 0.435    & 0.515 & 0.505  \\
advcl                  & 118                         & 0.430    & 0.512 & 0.518  \\
advmod                 & 197                         & 0.433    & 0.512 & 0.522  \\
amod                   & 82                          & 0.435    & 0.528 & 0.520  \\
appos                  & 109                         & 0.404    & 0.482 & 0.518  \\
aux                    & 147                         & 0.421    & 0.514 & 0.532  \\
aux:pass               & 147                         & 0.436    & 0.477 & 0.525  \\
\hline
\end{tabular}
\caption{Translation quality of Chinese sentences under different models according to dependency relations.}
\label{FullZhSyntaxRelations}
\end{table*}

\begin{table*}
\centering
\small
\begin{tabular}{l|c|ccc} 
\hline
\multicolumn{1}{c}{Bad Translations (Ru)} & \multicolumn{1}{c}{Sentences} & \multicolumn{1}{c}{Baseline} & \multicolumn{1}{c}{SGBC} & \multicolumn{1}{c}{SGBD}  \\ 
\hline
acl                    & 57                          & 0.708                    & 0.758                   & 0.749                      \\
acl:relcl              & 53                          & 0.706                    & 0.756                   & 0.747                      \\
advcl                  & 58                          & 0.695                    & 0.739                   & 0.752                      \\
advmod                 & 165                         & 0.704                    & 0.750                    & 0.747                      \\
amod                   & 228                         & 0.707                    & 0.753                   & 0.752                      \\
appos                  & 43                          & 0.695                    & 0.742                   & 0.740                       \\
aux                    & 10                          & 0.700                      & 0.764                   & 0.777                      \\
\hline
\end{tabular}
\caption{Translation quality of Russian sentences under different models according to dependency relations.}
\label{FullRuSyntaxRelations}
\end{table*}

\begin{table*}
\centering
\small
\begin{tabular}{l|c|ccc} 
\hline
\multicolumn{1}{c}{Bad Translations (De)} & \multicolumn{1}{c}{Sentences} & \multicolumn{1}{c}{Baseline} & \multicolumn{1}{c}{SGBC} & \multicolumn{1}{c}{SGBD}  \\ 
\hline
acl:relcl                        & 83                          & 0.506 & 0.578 & 0.582  \\
advcl                            & 56                          & 0.514 & 0.570 & 0.556  \\
advmod                           & 181                         & 0.506 & 0.573 & 0.582  \\
amod                             & 187                         & 0.507 & 0.567 & 0.571  \\
appos                            & 92                          & 0.500 & 0.556 & 0.565  \\
aux                              & 77                          & 0.520 & 0.586 & 0.597  \\
aux:pass                         & 62                          & 0.498 & 0.576 & 0.556  \\
\hline
\end{tabular}
\caption{Translation quality of German sentences under different models according to dependency relations.}
\label{FullDeSyntaxRelations}
\end{table*}

\begin{table*}
\centering
\small
\begin{tabular}{l|rr|rr|rr} 
\hline
             & \multicolumn{2}{c|}{Zh} & \multicolumn{2}{c|}{Ru} & \multicolumn{2}{c}{De}  \\
             & Samples & GAT           & Samples & GAT           & Samples & GAT           \\ 
\cline{2-7}
acl          & 120      & 0             & 256     & 0.396         & 20      & 0             \\
acl:relcl    & 448     & 0.917         & 160     & 0.359         & 271     & 0.612         \\
advcl        & 516     & 0.421        & 197     & 0.337         & 221     & 0.467         \\
advmod       & 1225    & 0.909         & 914     & 0.934         & 1120    & 0.932         \\
amod         & 419     & 0.927         & 1791    & 0.982         & 1101    & 0.913         \\
appos        & 248     & 0.405        & 121     & 0.440          & 265     & 0.523         \\
\hline
\end{tabular}
\caption{F1-score for GAT dependency prediction for three languages.}
\label{GATSyntaxPred}
\end{table*}

\begin{table*}
\centering
\small
\begin{tabular}{l|cccccccccccc} 
\hline
\multicolumn{13}{c}{\textbf{Baseline vs SGBC}}                                                                                       \\
\multicolumn{1}{l}{Zh Relations} & L1    & L2    & L3     & L4    & L5    & L6    & L7    & L8    & L9     & L10   & L11   & L12    \\ 
\hline
acl:relcl            & 0.891 & 0.733 & 0.877 & 0.239 & 0.452 & 0.656 & 0.506   & 0.712 & 0.587 & 0.623 & 0.442 & 0.424  \\
advcl                & 0.875 & 0.734 & 0.203  & 0.179 & 0.378 & 0.685 & 0.462 & 0.693 & 0.664  & 0.668 & 0.517 & 0.522  \\
advmod               & 0.856 & 0.794 & 0.878  & 0.292 & 0.528 & 0.781 & 0.576 & 0.733 & 0.638  & 0.697 & 0.514 & 0.512  \\
amod                 & 0.818 & 0.705 & 0.962  & 0.632 & 0.483 & 0.662 & 0.379 & 0.580  & 0.398  & 0.587  & 0.341 & 0.335  \\
appos                & 0.908 & 0.770  & 0.901  & 0.411 & 0.429 & 0.694 & 0.519 & 0.653 & 0.599  & 0.677 & 0.483 & 0.485  \\
aux                  & 0.873 & 0.803 & 0.954  & 0.149 & 0.600   & 0.760  & 0.551 & 0.718 & 0.614  & 0.683 & 0.476 & 0.441  \\
\hline
\end{tabular}
\caption{Comparison of the representation from BERT in the baseline and SGBC model when tested on Chinese sentences containing target dependency.}
\label{Zh-RSA}
\end{table*}

\begin{table*}
\centering
\small
\begin{tabular}{l|cccccccccccc} 
\hline
\multicolumn{13}{c}{\textbf{Baseline vs SGBD}}                                                                                                                                                                                                                                                                                         \\
\multicolumn{1}{l}{Zh Relations} & \multicolumn{1}{c}{L1} & \multicolumn{1}{c}{L2} & \multicolumn{1}{c}{L3} & \multicolumn{1}{c}{L4} & \multicolumn{1}{c}{L5} & \multicolumn{1}{c}{L6} & \multicolumn{1}{c}{L7} & \multicolumn{1}{c}{L8} & \multicolumn{1}{c}{L9} & \multicolumn{1}{c}{L10} & \multicolumn{1}{c}{L11} & \multicolumn{1}{c}{L12}  \\ 
\hline
acl:relcl            & 0.902                  & 0.726                  & 0.267                  & 0.237                  & 0.231                  & 0.574                  & 0.339                  & 0.554                  & 0.477                  & 0.554                   & 0.425                   & 0.444                    \\
advcl                & 0.893                  & 0.747                  & 0.178                  & 0.125                  & 0.251                  & 0.554                  & 0.425                  & 0.522                  & 0.454                  & 0.507                   & 0.386                   & 0.411                    \\
advmod               & 0.874                  & 0.768                  & 0.262                  & 0.260                  & 0.401                  & 0.664                  & 0.409                  & 0.492                  & 0.493                  & 0.581                   & 0.448                   & 0.548                    \\
\hline
\end{tabular}
\caption{Comparison of the representation from BERT in the baseline and SGBD model when tested on Chinese sentences containing target dependency.}
\label{Zh-RSA-2}
\end{table*}

\begin{table*}
\centering
\small
\begin{tabular}{l|cccccccccccc} 
\hline
\multicolumn{13}{c}{\textbf{Baseline vs SGBC}}                                                                                             \\
\multicolumn{1}{l}{Ru Relations} & L1     & L2    & L3    & L4     & L5     & L6    & L7    & L8    & L9    & L10   & L11   & L12    \\ 
\hline
acl                  & 0.824  & 0.424 & 0.392 & 0.625  & 0.555  & 0.738 & 0.646 & 0.618 & 0.571 & 0.644 & 0.641 & 0.559  \\
acl:relcl            & 0.617  & 0.309 & 0.310 & 0.454  & 0.412  & 0.640 & 0.519 & 0.635 & 0.576 & 0.553 & 0.506 & 0.475  \\
advcl                & 0.710  & 0.613 & 0.556 & 0.609  & 0.409  & 0.631 & 0.623 & 0.734 & 0.756 & 0.748 & 0.685 & 0.587  \\
advmod               & 0.877  & 0.608 & 0.428 & 0.651  & 0.618  & 0.764 & 0.711 & 0.723 & 0.721 & 0.746 & 0.734 & 0.618  \\
amod                 & 0.855  & 0.572 & 0.444 & 0.635  & 0.576  & 0.731 & 0.668 & 0.694 & 0.693 & 0.731 & 0.722 & 0.597  \\
appos                & 0.679  & 0.617 & 0.286 & 0.700  & 0.606  & 0.707 & 0.591 & 0.700 & 0.769 & 0.774 & 0.787 & 0.569  \\
\hline
\end{tabular}
\caption{Comparison of the representation from BERT in the baseline and SGBC model when tested on Russian sentences containing target dependency.}
\label{Ru-RSA}
\end{table*}

\begin{table*}
\centering
\small
\begin{tabular}{l|cccccccccccc} 
\hline
\multicolumn{13}{c}{\textbf{Baseline vs SGBD}}                                                                       \\
\multicolumn{1}{l}{Ru Relations} & L1    & L2    & L3    & L4    & L5    & L6    & L7    & L8    & L9    & L10   & L11   & L12    \\ 
\hline
acl                  & 0.918 & 0.416 & 0.296 & 0.617 & 0.501 & 0.541 & 0.611 & 0.562 & 0.573 & 0.752 & 0.779 & 0.627  \\
acl:relcl            & 0.505 & 0.299 & 0.292 & 0.484 & 0.402 & 0.474 & 0.606 & 0.643 & 0.628 & 0.739 & 0.744 & 0.651  \\
advcl                & 0.585 & 0.541 & 0.489 & 0.508 & 0.505 & 0.562 & 0.665 & 0.676 & 0.692 & 0.747 & 0.804 & 0.686  \\
advmod               & 0.931 & 0.442 & 0.509 & 0.608 & 0.613 & 0.646 & 0.731 & 0.720 & 0.710 & 0.830 & 0.846 & 0.666  \\
amod                 & 0.910 & 0.548 & 0.488 & 0.391 & 0.573 & 0.605 & 0.679 & 0.683 & 0.674 & 0.796 & 0.777 & 0.594  \\
appos                & 0.574 & 0.331 & 0.303 & 0.614 & 0.432 & 0.517 & 0.618 & 0.715 & 0.740 & 0.787 & 0.731 & 0.581  \\
\hline
\end{tabular}
\caption{Comparison of the representation from BERT in the baseline and SGBD model when tested on Russian sentences containing target dependency.}
\label{Ru-RSA-2}
\end{table*}

\begin{table*}
\centering
\small
\begin{tabular}{l|cccccccccccc} 
\hline
\multicolumn{13}{c}{\textbf{Baseline vs SGBC}}                                                                                                                                                                                                                                                                                            \\
De Relations & \multicolumn{1}{c}{L1} & \multicolumn{1}{c}{L2} & \multicolumn{1}{c}{L3} & \multicolumn{1}{c}{L4} & \multicolumn{1}{c}{L5} & \multicolumn{1}{c}{L6} & \multicolumn{1}{c}{L7} & \multicolumn{1}{c}{L8} & \multicolumn{1}{c}{L9} & \multicolumn{1}{c}{L10} & \multicolumn{1}{c}{L11} & \multicolumn{1}{c}{L12}  \\ 
\hline
acl:relcl    & 0.696                  & 0.776                  & 0.763                  & 0.690                  & 0.601                  & 0.604                  & 0.670                  & 0.627                  & 0.621                  & 0.629                   & 0.613                   & 0.629                    \\
advcl        & 0.640                  & 0.776                  & 0.781                  & 0.716                  & 0.645                  & 0.506                  & 0.632                  & 0.602                  & 0.572                  & 0.514                   & 0.527                   & 0.575                    \\
advmod       & 0.775                  & 0.819                  & 0.841                  & 0.800                  & 0.737                  & 0.733                  & 0.750                  & 0.793                  & 0.747                  & 0.790                   & 0.750                   & 0.748                    \\
amod         & 0.651                  & 0.739                  & 0.774                  & 0.721                  & 0.631                  & 0.662                  & 0.708                  & 0.645                  & 0.670                  & 0.641                   & 0.644                   & 0.663                    \\
\hline
\end{tabular}
\caption{Comparison of the representation from BERT in the baseline and SGBC model when tested on German sentences containing target dependency.}
\label{De-RSA}
\end{table*}

\begin{table*}
\centering
\small
\begin{tabular}{l|cccccccccccc} 
\hline
\multicolumn{13}{c}{\textbf{Baseline vs SGBD}}                                                                                                                                                                                                                                                                                                                \\
\multicolumn{1}{l}{De Relations} & \multicolumn{1}{c}{L1} & \multicolumn{1}{c}{L2} & \multicolumn{1}{c}{L3} & \multicolumn{1}{c}{L4} & \multicolumn{1}{c}{L5} & \multicolumn{1}{c}{L6} & \multicolumn{1}{c}{L7} & \multicolumn{1}{c}{L8} & \multicolumn{1}{c}{L9} & \multicolumn{1}{c}{L10} & \multicolumn{1}{c}{L11} & \multicolumn{1}{c}{L12}  \\ 
\hline
acl:relcl                        & 0.793                  & 0.740                  & 0.860                  & 0.831                  & 0.845                  & 0.883                  & 0.850                  & 0.828                  & 0.863                  & 0.801                   & 0.778                   & 0.689                    \\
advcl                            & 0.773                  & 0.720                  & 0.843                  & 0.815                  & 0.820                  & 0.894                  & 0.867                  & 0.856                  & 0.894                  & 0.842                   & 0.840                   & 0.747                    \\
advmod                           & 0.782                  & 0.796                  & 0.849                  & 0.832                  & 0.856                  & 0.859                  & 0.827                  & 0.774                  & 0.787                  & 0.783                   & 0.785                   & 0.794                    \\
\hline
\end{tabular}
\caption{Comparison of the representation from BERT in the baseline and SGBD model when tested on German sentences containing target dependency.}
\label{De-RSA-2}
\end{table*}

\end{document}